\documentclass{article}

\PassOptionsToPackage{numbers}{natbib}

\usepackage[final]{neurips_2019}

\usepackage[utf8]{inputenc} 
\usepackage[T1]{fontenc}    
\usepackage{hyperref}       
\usepackage{url}            
\usepackage{booktabs}       
\usepackage{amsfonts}       
\usepackage{nicefrac}       
\usepackage{microtype}      

\usepackage{import}         
\usepackage{xcolor}         
\usepackage{graphicx}
\usepackage{enumitem}       
\usepackage{amsmath}
\usepackage{multirow}
\usepackage{algpseudocode}

\hypersetup{
    colorlinks=true,
    citecolor=black,
    urlcolor=blue,
}

\newcommand{\para}[1]{{\bf #1\ \ }}

\title{LEAF: A Benchmark for Federated Settings}

\author{%
  Sebastian Caldas\thanks{Corresponding author. Correspondence to: \texttt{scaldas@cmu.edu}.}$^{*, 1}$, Sai Meher Karthik Duddu$^1$, Peter Wu$^1$, Tian Li$^1$, Jakub Kone{\v{c}}n{\'y}$^2$,\\
  \textbf{H. Brendan McMahan$^2$, Virginia Smith$^1$, Ameet Talwalkar$^{1, 3}$} \\
  \\
  $^1$ Carnegie Mellon University \\
  $^2$ Google \\
  $^3$ Determined AI \\
  \\
  \url{https://leaf.cmu.edu}
}

\begin{document}

\maketitle

\begin{abstract}
Modern federated networks, such as those comprised of wearable devices, mobile phones, or autonomous vehicles, generate massive amounts of data each day. This wealth of data can help to learn models that can improve the user experience on each device.
However, the scale and heterogeneity of federated data presents new challenges in research areas such as federated learning, meta-learning, and multi-task learning.
As the machine learning community begins to tackle these challenges, we are at a critical time to ensure that developments made in these areas are grounded with realistic benchmarks. 
To this end, we propose LEAF, a modular benchmarking framework for learning in federated settings.
LEAF includes a suite of open-source federated datasets, a rigorous evaluation framework, and a set of reference implementations, all geared towards capturing the obstacles and intricacies of practical federated environments.
\end{abstract}

\section{Introduction}

With data increasingly being generated on federated networks of remote devices,
there is growing interest in empowering on-device applications with models 
that make use of such data~\cite{mcmahan2016communication, McMahan:2017fl, smith2017federated, li2019federated, yang2019federated}. Learning on data generated in federated networks, however, introduces several new obstacles:

\para{Statistical:} Data is generated on each device in a heterogeneous manner, with each device associated with a different (though perhaps related) underlying data generating distribution. Moreover, the number of data points typically varies significantly across devices.

\para{Systems:} The number of devices in federated scenarios is typically order of magnitudes larger than the number of nodes in a typical distributed setting, such as datacenter computing. 
In addition, each device may have significant constraints in terms of storage, computational, and communication capacities. Furthermore, these capacities may also differ across devices due to variability in hardware, network connection, and power.
Thus, federated settings may suffer from communication bottlenecks that dwarf those encountered in traditional datacenter settings, and may require faster on-device inference. 

\para{Privacy and Security:} Finally, the sensitive nature of personally-generated data  
requires methods that operate on federated data to balance
privacy and security concerns with more traditional considerations such as 
statistical accuracy, scalability, and efficiency~\cite{mcmahan2018learning, bonawitz2017practical}.

Recent works have proposed diverse ways of dealing with these challenges, but many of these efforts fall short when it comes to their experimental evaluation. 
As an example, consider the federated learning paradigm, which focuses on training models directly on federated networks~\cite{mcmahan2016communication, smith2017federated, pihur2018differentially}.
Experimental works focused on federated learning broadly utilize three types of datasets, each with their own shortcoming:
(1) datasets that are commonly used and yet do not provide a realistic model of a federated scenario, e.g., artificial partitions of MNIST, MNIST-fashion or CIFAR-10~\cite{mcmahan2016communication, konevcny2016federated, geyer2017differentially, bagdasaryan2018backdoor, kamp2018efficient, ulm2018functional, wang2019adaptive};  
(2) realistic but proprietary federated datasets, e.g., data from an unnamed social network in~\cite{mcmahan2016communication}, crowdsourced voice commands in~\cite{leroy2019federated}, and proprietary data by Huawei in~\cite{chen2018federated}; and (3) realistic federated datasets that are derived from publicly available data, but which are not straightforward to reproduce,
e.g., FaceScrub in~\cite{melis2018inference}, Shakespeare in~\cite{mcmahan2016communication} and Reddit in~\cite{konevcny2016federated, mcmahan2018learning, bagdasaryan2018backdoor}. 

As a second example, consider meta-learning, a related learning paradigm proposed by~\cite{chen2018federated} and~\cite{khodak2019provable} as a way to tackle the statistical challenges of federated networks. 
The paradigm is indeed a natural fit for federated settings, as the heterogeneous devices can be interpreted as meta-learning tasks.
However, popular meta-learning benchmarks such as \emph{Omniglot}~\cite{lake2011one, finn2017model, vinyals2016matching, snell2017prototypical} and \emph{miniImageNet}~\cite{ravi2016optimization, finn2017model, vinyals2016matching, snell2017prototypical} focus on $k$-shot learning (i.e., all tasks have the same number of samples, each class has the same number of samples in each task, etc.) and thus fail to capture the real-world challenges that federated data would bring to meta-learning solutions.
In fact, all of the previously mentioned datasets could thus be categorized as the first type mentioned above (popular yet unrealistic for our purposes).

As a final example, consider multi-task learning (MTL). This paradigm is also amenable to federated settings~\cite{smith2017federated} but, contrary to realistic federated networks, is usually explored in regimes with small numbers of tasks and samples, e.g., the popular \emph{Landmine Detection} ~\cite{zhang2010learning, murugesan2017multi, xue2007multi, smith2017federated}, \emph{Computer Survey}~\cite{argyriou2008convex, agarwal2010learning, kumar2012learning} and \emph{Inner London Education Authority School}~\cite{murugesan2017multi, lee2016asymmetric, agarwal2010learning, argyriou2008convex, kumar2012learning} datasets have at most $200$ tasks each. 
We highlight that, while federated learning, meta-learning, and multi-task learning are the presented applications for LEAF, the framework in fact encompasses a wide range of potential learning settings, such as on-device learning or inference, transfer learning, life-long learning, and the development of personalized learning models.

Our work aims to bridge the gap between artificial datasets that are popular and accessible for benchmarking, and those that realistically capture the characteristics of a federated scenario but that, so far, have been either proprietary or difficult to process. 
Moreover, beyond  establishing a suite of federated datasets, we propose a clear methodology for evaluating methods and reproducing results. To this end, we present LEAF, a modular benchmarking framework geared towards learning in massively distributed federated networks of remote devices.

\section{LEAF}
\label{leaf}

\begin{figure}[ht]  
    \vskip 0.2in
    \begin{center}
    \centerline{\includegraphics[width=0.75\columnwidth]{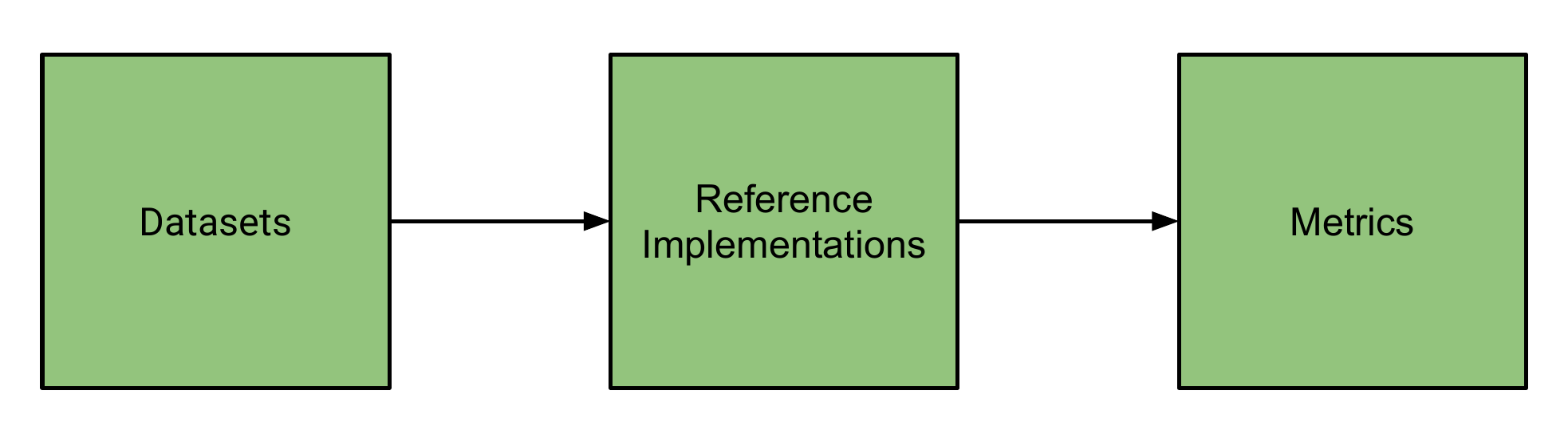}}
    \caption{LEAF modules. The “Datasets” module preprocesses the data and transforms it into a standardized format, which can integrate into an arbitrary ML pipeline. LEAF's “Reference Implementations” module is a growing repository of common methods used in the federated setting, with each implementation producing a log of various different statistical and systems metrics.
    Any log generated in an appropriate format can then be used to aggregate and analyze these metrics in various ways through LEAF's “Metrics” module.}
    \label{fig:modules}
    \end{center}
    \vskip -0.2in
\end{figure}

LEAF is an open-source benchmark for federated settings. 
\footnote{All code and documentation can be found at \url{https://github.com/TalwalkarLab/leaf/}.} It consists of (1) a suite of open-source datasets, (2) an array of statistical and systems metrics, and (3) a set of reference implementations. 
As shown in Figure~\ref{fig:modules}, LEAF's \emph{modular} design allows these three 
components
to be easily incorporated into diverse experimental pipelines. We proceed to detail LEAF's core components.

\para{Datasets:} We have curated a suite of realistic federated datasets for LEAF.
We focus on datasets where (1) the data has a natural keyed generation process (where each key refers to a particular device/user); (2) the data is generated from networks of thousands to millions of devices; and (3) the number of data points is skewed across devices. Currently, LEAF
consists of six datasets: 
\begin{itemize}[noitemsep, leftmargin=*]
    \item \textit{Federated Extended MNIST (FEMNIST)}, which is built by partitioning the data in Extended MNIST~\cite{lecun1998mnist, cohen2017emnist} based on the writer of the digit/character.
    \item \textit{Sentiment140}~\cite{go2009twitter}, an automatically generated sentiment analysis dataset that annotates tweets based on the emoticons present in them. Each device is a different twitter user.
    \item \textit{Shakespeare}, a dataset built from \textit{The Complete Works of William Shakespeare}~\cite{shakespeare, mcmahan2016communication}. Here, each speaking role in each play is considered a different device. 
    \item \textit{CelebA}, which partitions the Large-scale CelebFaces Attributes Dataset \footnote{The original CelebA data is hosted in \url{http://mmlab.ie.cuhk.edu.hk/projects/CelebA.html}}~\cite{liu2015faceattributes} by the celebrity on the picture. 
    \item \textit{Reddit}, where we preprocess comments posted on the social network on December 2017. 
    \item A \textit{Synthetic} dataset, which modifies the synthetic dataset presented in~\cite{li2019fair} to make it more challenging for current meta-learning methods. See Appendix~\ref{appendix:synth} for details.
\end{itemize}

We provide statistics on these datasets (except the Synthetic one, as these vary depending on the user's settings) in Table~\ref{table:datasets}. In LEAF, we provide all necessary pre-processing scripts for each dataset, as well as small/full versions for prototyping and final testing. Moving forward, we plan to add datasets from different domains (e.g. audio, video) and to increase the range of machine learning tasks (e.g. text to speech, translation, compression, etc.).

\begin{table*}[!t]
   \caption{Statistics of datasets in LEAF.}
   \label{table:datasets}
   \centering
   \small{
  \begin{tabular}{ccccc}
    \toprule
    \bf Dataset & \bf Number of devices & \bf Total samples & \multicolumn{2}{c}{\bf Samples per device} \\
    \cmidrule(l){4-5}
     &  &  & mean & stdev \\
    \midrule
    FEMNIST & $3,550$  & $805,263$ & $226.83$ & $88.94$ \\ \midrule
    Sent140 & $660,120$  & $1,600,498$ & $2.42$ & $4.71$ \\ \midrule
    Shakespeare & $1,129$  & $4,226,158$ & $3,743.28$ & $6,212.26$ \\ \midrule
    CelebA & $9,343$  & $200,288$ & $21.44$ & $7.63$ \\ \midrule
    Reddit & $1,660,820$  & $56,587,343$ & $34.07$ & $62.95$ \\
    \bottomrule
  \end{tabular}}
\end{table*}

\para{Metrics:} 
Rigorous evaluation metrics are required to appropriately assess how a learning solution behaves in federated scenarios. 
Currently, LEAF establishes an initial set of metrics chosen specifically for this purpose. 
For example, we introduce metrics that better capture the entire distribution of performance across devices: performance at the 10th, 50th and 90th percentiles and performance stratified by natural hierarchies in the data (e.g. “play” in the case of the Shakespeare dataset or “subreddit” for Reddit).
We also introduce metrics that account for the amount of computing resources needed from the edge devices in terms of number of FLOPS and number of bytes downloaded/uploaded.
Finally, LEAF also recognizes the importance of specifying how the accuracy is weighted across devices, e.g., whether every device is equally important, or every data point equally important (implying that power users/devices get preferential treatment).

\para{Reference implementations:}
In order to facilitate reproducibility, LEAF also contains a set of reference implementations of algorithms geared towards federated scenarios.
Currently, this set is limited to the federated learning paradigm, and in particular includes reference implementations of minibatch SGD, FedAvg~\cite{mcmahan2016communication} and Mocha~\cite{smith2017federated}.
Moving forward we aim to equip LEAF with implementations for additional methods and paradigms with the help of the broader research community.

\newpage

\section{LEAF in action}
\label{experiments}
\begin{figure}[t]
    \begin{center}
    \centerline{\includegraphics[width=0.8\columnwidth]{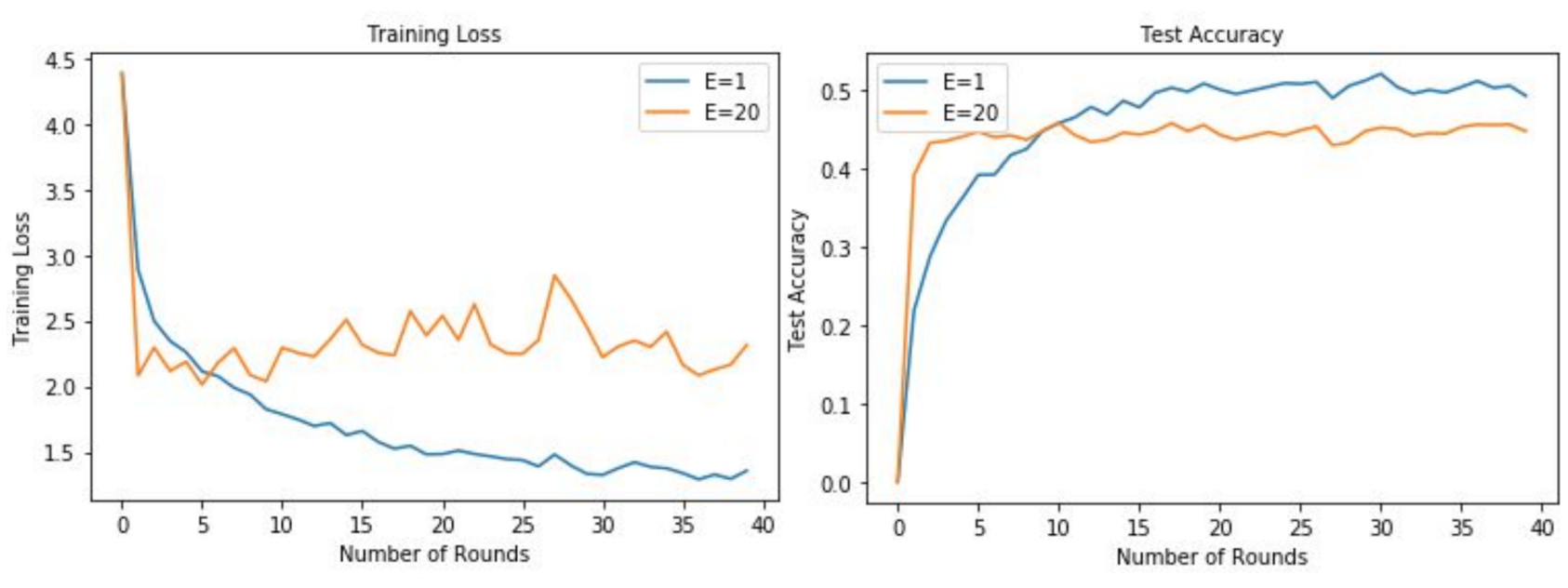}}
    \caption{Convergence behavior of FedAvg on a subsample of the Shakespeare dataset. We are able to achieve a per sample test accuracy comparable to the results obtained in~\cite{mcmahan2016communication}. We also qualitatively replicate the divergence in training loss that is observed for large numbers of local epochs ($E$).}
    \label{fig:shakespeare}
    \end{center}
\end{figure}

We now show a glimpse of LEAF in action. In particular, we highlight three of LEAF's characteristics\footnote{For experiment details, see Appendix~\ref{appendix:experiments}.}:

\para{LEAF enables reproducible science:} To demonstrate the reproducibility enabled via LEAF, we focus on qualitatively reproducing the results that~\cite{mcmahan2016communication} obtained on the Shakespeare dataset for a next character prediction task. In particular, it was noted that for this particular dataset, the FedAvg method surprisingly \textit{diverges} as the number of local epochs increases. This is therefore a critical setting to understand before deploying methods such as FedAvg. Results are shown in Figure~\ref{fig:shakespeare}, where we indeed see similar divergence behavior in terms of the training loss as we increase the number of epochs. 

\begin{figure}[t]
    \begin{center}
    \centerline{\includegraphics[width=0.8\columnwidth]{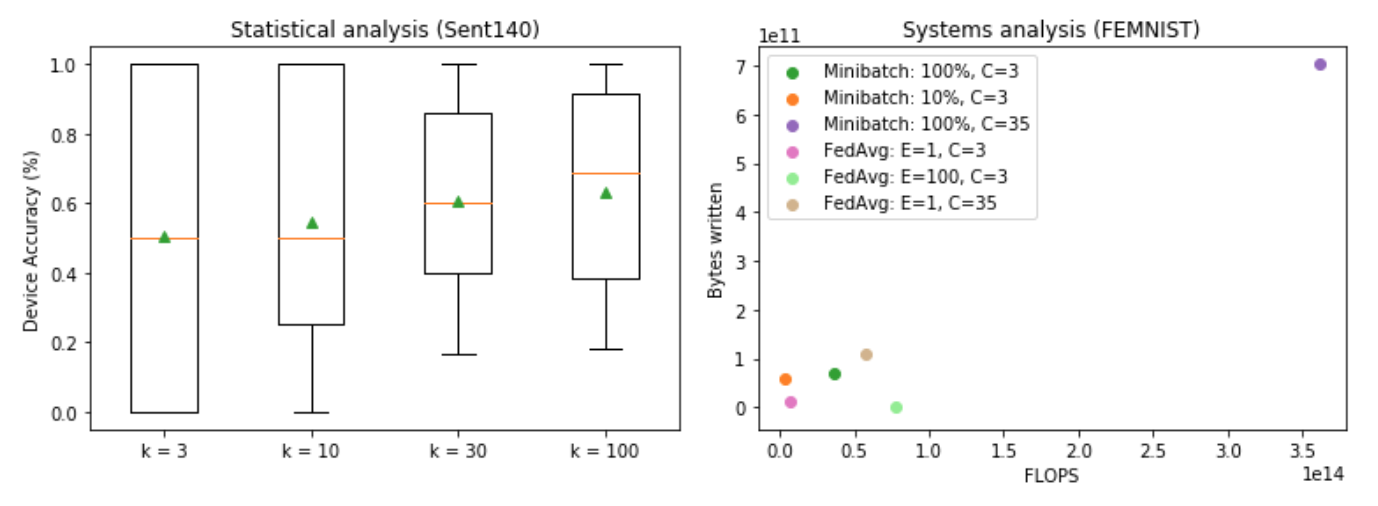}}
    \caption{Statistical and Systems analyses for Sent140 and FEMNIST. For Sent140: $k$ is the minimum number of samples per user. Orange lines represent the median device accuracy, green triangles represent the mean, boxes cover the 25th and 75th percentile, and whiskers cover the 10th to the 90th percentile. For FEMNIST: $C$ is the number of clients selected per round, and $E$ is the number of epochs each client trained locally for FedAvg. For minibatch SGD we report the percentage of data used per client.}
    \label{fig:sys_stat}
    \end{center}
    \vspace{-5mm}
\end{figure}

\para{LEAF provides granular metrics:} As illustrated in Figure~\ref{fig:sys_stat}, our proposed systems and statistical metrics are important to consider when serving multiple clients simultaneously. 
For statistical metrics, we show the effect of varying the minimum number of samples per user in Sentiment140 (which we denote as $k$). 
We see that, while median performance degrades only slightly with data-deficient users (i.e., $k=3$), the 25th percentile degrades dramatically.  
Meanwhile, for systems metrics, we run minibatch SGD and FedAvg for FEMNIST and calculate the systems budget needed to reach a per sample accuracy threshold of $0.75$. 
We characterize the budget in terms of total number of FLOPS across all devices and total number of bytes uploaded to network. 
Our results demonstrate the improved systems profile of FedAvg when it comes to the communication vs. local computation trade-off, though we note that in general methods may vary across these two dimensions. 

\para{LEAF is modular:} To demonstrate LEAF's modularity, we incorporate its “Datasets” module into three new experimental pipelines: one that trains purely local models for each device (on CelebA and our Synthetic dataset), one that disregards the natural partition between devices, i.e., it mixes all the data (on Reddit), and one in which we use the popular meta-learning method \emph{Reptile}~\cite{nichol2018first} (on FEMNIST). Results for these experiments are presented in Table~\ref{table:pipelines}. These particular pipelines shed light on how different modeling approaches may be more or less appropriate for different federated datasets~\cite{jiang2019improving, khodak2019provable}.

\begin{table*}[!t]
   \caption{Demonstration of LEAF's modularity. We incorporate LEAF's datasets into new experimental pipelines (beyond FedAvg) and report the resulting sample test accuracies.}
   \label{table:pipelines}
   \centering
   \small{
  \begin{tabular}{cccc}
    \toprule
    \bf Dataset & \bf FedAvg (baseline) & \multicolumn{2}{c}{\bf Additional Pipeline} \\
    \cmidrule(l){3-4}
     &  & description & accuracy \\
    \midrule
    CelebA & $89.46\%$  &  \multirow{2}{*}{Local Models} & $65.29\%$ \\ \cmidrule(lr){1-2} \cmidrule(lr){4-4}
    Synthetic & $71.89\%$ &  & $87.34\%$ \\ \midrule
    Reddit & $13.35\%$  & Global IID model & $12.60\%$ \\ \midrule
    FEMNIST & $74.72\%$ & Reptile  & $80.24\%$ \\
    \bottomrule
  \end{tabular}}
\end{table*}

\section{Conclusions}
We present LEAF, a modular framework for learning in federated settings, or ecosystems marked by massively distributed networks of devices. 
Learning paradigms applicable in such settings include federated learning, meta-learning, multi-task learning, and on-device learning.

LEAF will allow researchers and practitioners in domains such as federated learning, meta-learning, and multi-task learning to reason about new proposed solutions under more realistic assumptions than previous benchmarks. 
We intend to keep LEAF up to date with new datasets, metrics and open-source solutions in order to foster informed and grounded progress in this field.

\subsubsection*{Acknowledgements}

This work was supported in part by DARPA FA875017C0141, the National Science Foundation grants IIS1705121 and IIS1838017, an Okawa Grant, a Google Faculty Award, an Amazon Web Services Award, a JP Morgan A.I. Research Faculty Award, a Carnegie Bosch Institute Research Award, and the CONIX Research Center, one of six centers in JUMP, a Semiconductor Research Corporation (SRC) program sponsored by DARPA. Any opinions, findings and conclusions or recommendations expressed in this material are those of the author(s) and do not necessarily reflect the views of DARPA, the National Science Foundation, or any other funding agency.

\bibliography{bibliography}

\begin{thebibliography}{10}

\bibitem{agarwal2010learning}
Arvind Agarwal, Samuel Gerber, and Hal Daume.
\newblock Learning multiple tasks using manifold regularization.
\newblock In {\em Advances in neural information processing systems}, 2010.

\bibitem{argyriou2008convex}
Andreas Argyriou, Theodoros Evgeniou, and Massimiliano Pontil.
\newblock Convex multi-task feature learning.
\newblock {\em Machine Learning}, 73(3):243--272, 2008.

\bibitem{bagdasaryan2018backdoor}
Eugene Bagdasaryan, Andreas Veit, Yiqing Hua, Deborah Estrin, and Vitaly
  Shmatikov.
\newblock How to backdoor federated learning.
\newblock {\em arXiv preprint arXiv:1807.00459}, 2018.

\bibitem{bonawitz2017practical}
Keith Bonawitz, Vladimir Ivanov, Ben Kreuter, Antonio Marcedone, H~Brendan
  McMahan, Sarvar Patel, Daniel Ramage, Aaron Segal, and Karn Seth.
\newblock Practical secure aggregation for privacy-preserving machine learning.
\newblock In {\em ACM SIGSAC Conference on Computer and Communications
  Security}, 2017.

\bibitem{chen2018federated}
Fei Chen, Zhenhua Dong, Zhenguo Li, and Xiuqiang He.
\newblock Federated meta-learning for recommendation.
\newblock {\em arXiv preprint arXiv:1802.07876}, 2018.

\bibitem{cohen2017emnist}
Gregory Cohen, Saeed Afshar, Jonathan Tapson, and Andr{\'e} van Schaik.
\newblock {EMNIST}: an extension of {MNIST} to handwritten letters.
\newblock {\em arXiv preprint arXiv:1702.05373}, 2017.

\bibitem{finn2017model}
Chelsea Finn, Pieter Abbeel, and Sergey Levine.
\newblock Model-agnostic meta-learning for fast adaptation of deep networks.
\newblock In {\em International Conference on Machine Learning}, 2017.

\bibitem{geyer2017differentially}
Robin~C Geyer, Tassilo Klein, and Moin Nabi.
\newblock Differentially private federated learning: A client level
  perspective.
\newblock {\em arXiv preprint arXiv:1712.07557}, 2017.

\bibitem{go2009twitter}
Alec Go, Richa Bhayani, and Lei Huang.
\newblock Twitter sentiment classification using distant supervision.
\newblock {\em Project Report, Stanford}, 2009.

\bibitem{jiang2019improving}
Yihan Jiang, Jakub Kone{\v{c}}n{\`y}, Keith Rush, and Sreeram Kannan.
\newblock Improving federated learning personalization via model agnostic meta
  learning.
\newblock {\em arXiv preprint arXiv:1909.12488}, 2019.

\bibitem{kamp2018efficient}
Michael Kamp, Linara Adilova, Joachim Sicking, Fabian H{\"u}ger, Peter
  Schlicht, Tim Wirtz, and Stefan Wrobel.
\newblock Efficient decentralized deep learning by dynamic model averaging.
\newblock In {\em Joint European Conference on Machine Learning and Knowledge
  Discovery in Databases}, 2018.

\bibitem{khodak2019provable}
Mikhail Khodak, Maria Florina-Balcan, and Ameet Talwalkar.
\newblock Adaptive gradient-based meta-learning methods.
\newblock {\em Advances in Neural Information Processing Systems}, 2019.

\bibitem{konevcny2016federated}
Jakub Kone{\v{c}}n{\`y}, H~Brendan McMahan, Felix~X Yu, Peter Richt{\'a}rik,
  Ananda~Theertha Suresh, and Dave Bacon.
\newblock Federated learning: Strategies for improving communication
  efficiency.
\newblock {\em arXiv preprint arXiv:1610.05492}, 2016.

\bibitem{kumar2012learning}
Abhishek Kumar and Hal Daume~III.
\newblock Learning task grouping and overlap in multi-task learning.
\newblock {\em arXiv preprint arXiv:1206.6417}, 2012.

\bibitem{lake2011one}
Brenden Lake, Ruslan Salakhutdinov, Jason Gross, and Joshua Tenenbaum.
\newblock One shot learning of simple visual concepts.
\newblock In {\em Annual Meeting of the Cognitive Science Society}, 2011.

\bibitem{lecun1998mnist}
Yann LeCun.
\newblock The {MNIST} database of handwritten digits.
\newblock {\em \url{http://yann.lecun.com/exdb/mnist/}}, 1998.

\bibitem{lee2016asymmetric}
Giwoong Lee, Eunho Yang, and Sung Hwang.
\newblock Asymmetric multi-task learning based on task relatedness and loss.
\newblock In {\em International Conference on Machine Learning}, 2016.

\bibitem{leroy2019federated}
David Leroy, Alice Coucke, Thibaut Lavril, Thibault Gisselbrecht, and Joseph
  Dureau.
\newblock Federated learning for keyword spotting.
\newblock In {\em IEEE International Conference on Acoustics, Speech and Signal
  Processing}, 2019.

\bibitem{li2019federated}
Tian Li, Anit~Kumar Sahu, Ameet Talwalkar, and Virginia Smith.
\newblock Federated learning: Challenges, methods, and future directions.
\newblock {\em arXiv preprint arXiv:1908.07873}, 2019.

\bibitem{li2019fair}
Tian Li, Maziar Sanjabi, and Virginia Smith.
\newblock Fair resource allocation in federated learning.
\newblock {\em arXiv preprint arXiv:1905.10497}, 2019.

\bibitem{liu2015faceattributes}
Ziwei Liu, Ping Luo, Xiaogang Wang, and Xiaoou Tang.
\newblock Deep learning face attributes in the wild.
\newblock In {\em International Conference on Computer Vision}, 2015.

\bibitem{mcmahan2016communication}
H~Brendan McMahan, Eider Moore, Daniel Ramage, Seth Hampson, and Blaise~Aguera
  y~Arcas.
\newblock Communication-efficient learning of deep networks from decentralized
  data.
\newblock In {\em Artificial Intelligence and Statistics}, 2017.

\bibitem{McMahan:2017fl}
H~Brendan McMahan and Daniel Ramage.
\newblock
  \href{http://www.googblogs.com/federated-learning-collaborative-machine-learning-without-centralized-training-data/}{Federated
  Learning: Collaborative Machine Learning without Centralized Training Data}.
\newblock {\em googblogs.com}, 2017.

\bibitem{mcmahan2018learning}
H~Brendan McMahan, Daniel Ramage, Kunal Talwar, and Li~Zhang.
\newblock Learning differentially private recurrent language models.
\newblock In {\em International Conference on Learning Representations}, 2018.

\bibitem{melis2018inference}
Luca Melis, Congzheng Song, Emiliano De~Cristofaro, and Vitaly Shmatikov.
\newblock Inference attacks against collaborative learning.
\newblock {\em arXiv preprint arXiv:1805.04049}, 2018.

\bibitem{murugesan2017multi}
Keerthiram Murugesan and Jaime Carbonell.
\newblock Multi-task multiple kernel relationship learning.
\newblock In {\em SIAM International Conference on Data Mining}, 2017.

\bibitem{nichol2018first}
Alex Nichol, Joshua Achiam, and John Schulman.
\newblock On first-order meta-learning algorithms.
\newblock {\em arXiv preprint arXiv:1803.02999}, 2018.

\bibitem{pihur2018differentially}
Vasyl Pihur, Aleksandra Korolova, Frederick Liu, Subhash Sankuratripati, Moti
  Yung, Dachuan Huang, and Ruogu Zeng.
\newblock Differentially-private "draw and discard" machine learning.
\newblock {\em arXiv preprint arXiv:1807.04369}, 2018.

\bibitem{ravi2016optimization}
Sachin Ravi and Hugo Larochelle.
\newblock Optimization as a model for few-shot learning.
\newblock {\em \url{https://openreview.net/pdf?id=rJY0-Kcll}}, 2016.

\bibitem{smith2017federated}
Virginia Smith, Chao-Kai Chiang, Maziar Sanjabi, and Ameet~S Talwalkar.
\newblock Federated multi-task learning.
\newblock In {\em Advances in Neural Information Processing Systems}, 2017.

\bibitem{snell2017prototypical}
Jake Snell, Kevin Swersky, and Richard Zemel.
\newblock Prototypical networks for few-shot learning.
\newblock In {\em Advances in Neural Information Processing Systems}, 2017.

\bibitem{ulm2018functional}
Gregor Ulm, Emil Gustavsson, and Mats Jirstrand.
\newblock Functional federated learning in erlang (ffl-erl).
\newblock In {\em International Workshop on Functional and Constraint Logic
  Programming}, 2018.

\bibitem{vinyals2016matching}
Oriol Vinyals, Charles Blundell, Tim Lillicrap, Daan Wierstra, et~al.
\newblock Matching networks for one shot learning.
\newblock In {\em Advances in Neural Information Processing Systems}, 2016.

\bibitem{wang2019adaptive}
Shiqiang Wang, Tiffany Tuor, Theodoros Salonidis, Kin~K Leung, Christian
  Makaya, Ting He, and Kevin Chan.
\newblock Adaptive federated learning in resource constrained edge computing
  systems.
\newblock {\em IEEE Journal on Selected Areas in Communications}, 2019.

\bibitem{shakespeare}
{William Shakespeare. The Complete Works of William Shakespeare}.
\newblock Publicly available at \url{//www.gutenberg.org/ebooks/100}.

\bibitem{xue2007multi}
Ya~Xue, Xuejun Liao, Lawrence Carin, and Balaji Krishnapuram.
\newblock Multi-task learning for classification with dirichlet process priors.
\newblock {\em Journal of Machine Learning Research}, 8(Jan):35--63, 2007.

\bibitem{yang2019federated}
Qiang Yang, Yang Liu, Tianjian Chen, and Yongxin Tong.
\newblock Federated machine learning: Concept and applications.
\newblock {\em ACM Transactions on Intelligent Systems and Technology}, 2019.

\bibitem{zhang2010learning}
Yi~Zhang and Jeff~G Schneider.
\newblock Learning multiple tasks with a sparse matrix-normal penalty.
\newblock In {\em Advances in Neural Information Processing Systems}, 2010.

\end{thebibliography}
\bibliographystyle{plain}

\newpage

\appendix
\section{Synthetic Dataset}
\label{appendix:synth}
Our synthetic dataset (introduced in Section~\ref{leaf}) is inspired by the one presented in~\cite{li2019fair}, but has possible additional heterogeneity designed to make current meta-learning methods (such as \emph{Reptile}~\cite{nichol2018first}) fail. The high-level goal is to create tasks whose true models are (1) task-dependant, and (2) clustered around more than just one center. 

To start, the user must input the desired number of devices $T \geq 1$ and a vector $(p_1, \dots, p_k)$ such that $p_j > 0, j \in 1, \dots, k$ and $\sum_{j=1}^k p_j = 1$. As preparation to generate the tasks:
\begin{enumerate}[noitemsep, leftmargin=*]
    \item Sample cluster means $\mu_j \in \mathbb{R}^s, j \in 1, \dots, k$. To do this, draw $\mu_j \sim N(B_j, I), B_j \sim N(0, I)$. 
    \item Draw matrix $Q \in \mathbb{R}^{d+1 \times s}$ by sampling $Q \sim N(0, I)$.
    \item Create diagonal matrix $\Sigma$ such that $\Sigma_{i,i} = i^{-1.2}$.
\end{enumerate}

Now, for each task $t \in 1, \dots, T$:
\begin{enumerate}[noitemsep, leftmargin=*]
    \item Sample a cluster center $\mu_t$ according to the input probabilities $(p_1, \dots, p_k)$.
    \item Draw $u_t \sim N(\mu_t, I)$ and set $w_t = Q u_t, w_t \in \mathbb{R}^{d+1}$.
    \item Now, draw $m_t$ from a log-normal distribution with mean $3$ and sigma $2$. We then set the number of samples $n_t = \min(m_t + 5, 1000)$ (to put a lower and an upper bound on the number of samples per task).
    \item Sample $v_t \sim N(C_t, I), C_t \sim N(0, I)$. 
    \item Now, for $i \in 1, \dots, n_t$, sample $x_t^i \in \mathbb{R}^d$ by drawing $x_t^i \sim N(v_t, \Sigma)$.
    \item Finally, set $y_t^i = \arg \max(\text{sigmoid}(w_t x_t^i + N(0, 0.1 \cdot I)))$ after adding the necessary padding to $x_t^i$ to account for the intercept. 
\end{enumerate}

\section{Experiment Details}
\label{appendix:experiments}
In this section, we provide details for the experiments presented in Section~\ref{experiments}. 

\para{Shakespeare convergence.} For the experiment presented in Figure~\ref{fig:shakespeare}, we subsample $118$ devices (around $5\%$ of the total) in our Shakespeare data. Our model first maps each character to an embedding of dimension $8$ before passing it through an LSTM of two layers of $256$ units each. The LSTM emits an output embedding, which is scored against all items of the vocabulary via dot product followed by a softmax. We use a sequence length of $80$ for the LSTM. We evaluate using \texttt{AccuracyTop1}. We use a learning rate of $0.8$ and $10$ devices per round for all experiments.

\para{Statistical and systems analyses.} For all the Sent140 experiments presented in Figure~\ref{fig:sys_stat}, we use a bag of words model with logistic regression, and a learning rate of $3 \cdot 10^{-4}$. For the FEMNIST experiments in the same figure, we subsample $5\%$ of the data, and use a model with two convolutional layers followed by pooling, and a final dense layer with 2048 units. We use a learning rate of $4 \cdot 10^{-3}$ for FedAvg and of $6 \cdot 10^{-2}$ for minibatch SGD.

\para{Additional pipelines.} For the experiments presented in Table~\ref{table:pipelines} we use a split of $60\%$ training, $20\%$ validation and $20\%$ test per user, and report results on the test set. The hyperparameters that vary per experiment are the following:
\begin{itemize}[noitemsep, leftmargin=*]
    \item For the CelebA experiments, we use $10\%$ of the total clients and the same model we described above for FEMNIST. For the local models, each device explored learning rates in $[0.1, 0.01, 0.001, 0.0001]$. The FedAvg model uses 10 clients per round for 100 rounds, training locally for one epoch with a batch size of 5, and a best learning rate of $0.001$. Both results are averaged over 5 runs.
    \item For the experiments with the Synthetic dataset, we use $1,000$ devices, only one cluster, $60$ features and $5$ classes. Our model is a perceptron with sigmoid activations. For the local models, each device explored learning rates in $[10^{-3}, 10^{-2}, 10^{-1}, 1, 10, 10^{2}, 10^{3}]$. The FedAvg model used 10 clients per round for 100 rounds, trained locally for one epoch with a batch size of 5, and found a best learning rate of $0.1$. 
    \item For the Reddit experiments, we use 819 devices and a model similar to the one we described for Shakespeare. The main differences are: the size of the embedding is now $200$, and we build the vocabulary from the tokens in the training set with a fixed length of $10,000$. We use a sequence length of $10$, evaluate using \texttt{AccuracyTop1} and consider all predictions of the unknown and padding tokens as incorrect. For the global iid model, we train for 3 epochs over all the devices' data using a learning rate of $4 \cdot \sqrt{2}$. For FedAvg, we use 10 clients per round for 100 rounds, training locally for one epoch using a batch size of 5. We use a learning rate of $8$. Both results are averaged over 5 runs.
    \item For the FEMNIST experiments we use the same model as described before and run each algorithm for $1,000$ rounds, use $5$ clients per round, a local learning rate of $10^{-3}$, a training mini-batch size of $10$ for $5$ mini-batches, and evaluate on an unseen set of test devices. Furthermore, for \emph{Reptile} we use a linearly decaying meta-learning rate that goes from $2$ to $0$, and evaluate by fine-tuning each test device for $50$ mini-batches of size $5$.
\end{itemize}

\end{document}